\documentclass{article}

\usepackage{iclr2019_conference,times}

\usepackage{amsmath,amsfonts,bm}

\def\eqref#1{equation~\ref{#1}}

\def\1{\bm{1}}

\def\mW{{\bm{W}}}

\DeclareMathAlphabet{\mathsfit}{\encodingdefault}{\sfdefault}{m}{sl}
\SetMathAlphabet{\mathsfit}{bold}{\encodingdefault}{\sfdefault}{bx}{n}
\newcommand{\tens}[1]{\bm{\mathsfit{#1}}}
\def\tA{{\tens{A}}}
\def\tB{{\tens{B}}}
\def\tC{{\tens{C}}}

\def\tM{{\tens{M}}}

\def\tW{{\tens{W}}}
\def\tX{{\tens{X}}}
\def\tY{{\tens{Y}}}

\newcommand{\R}{\mathbb{R}}

\usepackage{graphicx}
\graphicspath{ {./eps/} }

\usepackage{hyperref}
\hypersetup{
colorlinks=true,
linkcolor=blue,
filecolor=magenta,
urlcolor=cyan,
}

\title{Feedbackward Decoding for Semantic Segmentation}

\author{Beinan Wang, John Glossner, Daniel Iancu \\
Optimum Semiconductor Technologies \\
Tarrytown, NY 10591, USA \\
\texttt{\{bwang,jglossner,diancu\}@optimumsemi.com} 
\AND
Georgi N. Gaydadjiev \\
Department of Computer Science and Engineering \\
Delft University of Technology \\
Delft, the Netherlands \\
\texttt{\{g.n.gaydadjiev\}@tudelft.nl}
}

\iclrfinalcopy

\begin{document}

\maketitle

\begin{abstract}
We propose a novel approach for semantic segmentation that uses an encoder in the reverse direction to decode. Many semantic segmentation networks adopt a feedforward encoder-decoder architecture. Typically, an input is first downsampled by the encoder to extract high-level semantic features and continues to be fed forward through the decoder module to recover low-level spatial clues. Our method works in an alternative direction that lets information flow backward from the last layer of the encoder towards the first. The encoder performs encoding in the forward pass and the same network performs decoding in the backward pass. Therefore, the encoder itself is also the decoder. Compared to conventional encoder-decoder architectures, ours doesn't require additional layers for decoding and further reuses the encoder weights thereby reducing the total number of parameters required for processing.  We show by using only the 13 convolutional layers from VGG-16 plus one tiny classification layer, our model significantly outperforms other frequently cited models that are also adapted from VGG-16. On the Cityscapes semantic segmentation benchmark, our model uses 50.0\% less parameters than SegNet and achieves a 18.1\% higher "IoU class" score; it uses 28.3\% less parameters than DeepLab LargeFOV and the achieved "IoU class" score is 3.9\% higher; it uses 89.1\% less parameters than FCN-8s and the achieved "IoU class" score is 3.1\% higher. Our code will be publicly available on Github later.
\end{abstract}

\section{Introduction}
Various semantic segmentation networks, either earlier networks like U-Net \citep{DBLP:journals/corr/RonnebergerFB15}, or state of the art models such as DeepLabv3+ \citep{DBLP:journals/corr/abs-1802-02611} employ an encoder-decoder architecture. This type of network architectures first downsample data through an encoder to obtain coarse abstract features. Then features are upsampled using a decoder module to recover fine spatial details. Feature maps with the same dimensions from the encoder and the decoder modules are often fused to complement each other.

The encoder-decoder architecture has shown its effectiveness on a wide range of tasks from medical image segmentation \citep{10.1093/bioinformatics/btu080} to street scene parsing \citep{BROSTOW200988,MVD2017}. However, the decoder and the optional bridge between the encoder and the decoder require a considerable number of parameters. Moreover, unlike the encoder that is often pretrained on the ImageNet dataset \citep{DBLP:journals/corr/RussakovskyDSKSMHKKBBF14}, the decoder needs to be trained from scratch, thus leading to an information asymmetry between the encoder and the decoder.

Given the drawbacks of attaching a decoder to an encoder, we evaluate  if semantic segmentation with an encoder and no dedicated decoding layers can be employed without degradation. Our technique does not modify the architecture of the encoder, and therefore can be used with any base network - e.g., VGG \citep{Simonyan14c}, ResNet \citep{DBLP:journals/corr/HeZRS15} and Xception \citep{DBLP:journals/corr/Chollet16a}.

This paper is organized as follows, in Section \ref{sec:related_work} we discuss semantic segmentation related work. In Section \ref{sec:methods} we describe the theory behind feedbackward decoding. In Section \ref{sec:experiments} we describe quantitative experiments performed using feedbackward decoding. In Section \ref{sec:results} we compare our approach with other methods of transforming low resolution feature maps to high resolution score maps. In Section \ref{sec:discussion} we discuss optimization of weight matrices. Finally, in Section \ref{sec:conclusions} we conclude with observations and summarize our findings.

\section{Related Work}\label{sec:related_work}
The objective of semantic segmentation is to find a mapping from an input image $\displaystyle \tX \in \R^{h \times w \times c}$ to a segmentation mask $\displaystyle \tM \in \R^{h \times w}$, where $\displaystyle h$ is the height of the input image, $\displaystyle w$ the width and $\displaystyle c$ the number of channels. This problem is often solved by first mapping $\displaystyle \tX$ to $\displaystyle k$ score maps or collectively as a score tensor $\displaystyle \tY \in \R^{h \times w \times k}$, where $\displaystyle k$ is the number of possible classes in the scene. Each pixel is then labeled with the class that has the highest score. 

Traditional methods use a variety of techniques. Section 3 of \cite{DBLP:journals/corr/abs-1809-10198} provides a detailed overview of those for the interested readers. Most of the traditional techniques and methods became obsolete once deep learning based algorithms were introduced. Among deep learning based algorithms, the Fully Convolutional Network \citep{DBLP:journals/corr/LongSD14} is one of the most popular frameworks due to its ability to be trained end-to-end and the flexibility to process inputs of arbitrary resolutions.

Fully convolutional networks usually borrow a subset of layers from state of the art classifiers as their feature extractors, on top of which task-specific layers can be added. Because those classifiers are designed to progressively increase the number of feature maps, they often have to gradually decrease the size of feature maps to compensate for the increased resource requirements. Consequently, spatial information is lost as data flows downstream. This is not a problem for classification-only tasks as we only care about the dominating class in the image but semantic segmentation requires a pixel-to-pixel mapping from image to class. Therefore, there exists a contention between acquiring semantic features and preserving spatial information.

To resolve this contention, numerous solutions have been proposed. \cite{DBLP:journals/corr/LongSD14} applies transposed convolution with fractional input strides. Despite its name, transposed convolution does not really use the transpose of existing convolutional weights. Instead, the weights of transposed convolutional layers are initialized to perform bilinear interpolation, and then updated through training. When used with fractional input strides, transposed convolution scales up spatial dimensions by factors that are equal to the inverses of the input strides. Fractionally-strided convolution combines resizing and convolution into a single step but the convolutional filters may be poorly utilized due to the sparsity of the 0-inserted feature maps. Section 4.6 of \cite{Dumoulin2016AGT} gives a good illustration of this problem. In addition, transposed convolutions are further burdened when filter dimensions are not divisible by the input strides \citep{odena2016deconvolution}.

Networks such as the ones in \cite{Chen2014SemanticIS,DBLP:journals/corr/ChenPK0Y16,DBLP:journals/corr/ChenPSA17} and \cite{Yu2015MultiScaleCA} attempt to prevent reduced resolution by limiting the number of downsampling operations. However, this only relieves the symptom but does not cure the cause because computation at full resolution is often infeasible. As a result, log probabilities are usually predicted at a reasonable scale and then interpolated to full resolution. Alternatively,  in \cite{DBLP:journals/corr/BadrinarayananK15} max unpooling is used in place of interpolation.

Interpolation or unpooling alone is further complicated in that the produced score maps are often too coarse to capture sharp details. To refine the results, models like SegNet \citep{DBLP:journals/corr/BadrinarayananK15}, U-Net \citep{DBLP:journals/corr/RonnebergerFB15}, RefineNet \citep{DBLP:journals/corr/LinMS016} and DeepLabv3+ \citep{DBLP:journals/corr/abs-1802-02611} chain upsampling with convolution to achieve learnable upsampling. These models typically employ an encoder-decoder architecture in which the encoder performs convolutions followed by downsampling and the decoder performs upsampling followed by convolutions. Currently, this architecture is one of the most effective ways to approach semantic segmentation and noteably DeepLabv3+ achieves state of the art results on many benchmarks.

In spite of the encoder-decoder architecture being well known, to the best of the authors' knowledge, there has yet to be a publication that discusses how to reuse existing encoder parameters to decode. In a decoder module, not only are features mapped from smaller spatial dimensions to larger ones, but they also have to be projected from a higher channel dimension to a lower one. This is currently achieved by adding more convolutional layers. However, unlike their pretrained encoding counterparts, the new decoding convolutional filters do not have any previous experience; hence they are considerably harder to train and often become the performance bottleneck in the system. Our proposed method solves both problems with a single solution that allows convolutional filters in the downsampling and channel expanding path to carry over their knowledge and use it to guide semantic segmentation in the upsampling and channel contracting path.

\section{Methods}\label{sec:methods}
Feedbackward decoding is influenced by multitask learning that is present in many machine learning applications. In \cite{DBLP:journals/corr/RedmonDGF15,DBLP:journals/corr/RedmonF16,DBLP:journals/corr/abs-1804-02767}, the same layers are used for both classification and bounding box prediction; Multitask Question Answering Network \citep{DBLP:journals/corr/abs-1806-08730} is trained to perform a variety of complex natural language tasks including machine translation, summarization, sentiment analysis, etc. These works differ greatly in their objectives but they share a common design point: layers in a neural network are capable of accomplishing different goals. It turns out, multitask learning is not only possible, but also improves generalization because the shared parameters are more constrained towards good values (\citealp[P.~244]{Goodfellow-et-al-2016}).

Inspired by the possibility of a neural network to perform multiple tasks, our method trains a network to do both encoding and decoding. Our goal is to construct a decoder using the existing encoding layers. Consider a convolutional layer L = Conv2D($\displaystyle c_1$, $\displaystyle c_2$, ($\displaystyle m$, $\displaystyle n$)) with unit strides, "same" padding \citep{Dumoulin2016AGT} and no bias in an encoder module, where $\displaystyle c_1$ is the number of input channels, $\displaystyle c_2$ the number of output channels, $\displaystyle m$ the filter height and $\displaystyle n$ the filter width. Given an input tensor $\displaystyle \tA \in \R^{h \times w \times c_1}$, L produces an output tensor $\displaystyle \tB \in \R^{h \times w \times c_2}$. If desired $\displaystyle \tB$ can be further downsampled to a tensor $\displaystyle \tC \in \R^{\frac{h}{s} \times \frac{w}{t} \times c_2}$ through pooling with non-unit strides $\displaystyle s$ and $\displaystyle t$. Alternatively, pooling can be replaced with setting the strides of L to be $\displaystyle s$ and $\displaystyle t$.

In order for the decoder to transform $\displaystyle \tC$ back to a tensor $\displaystyle \tA' \in \R^{h \times w \times c_1}$ that has the same dimensions as $\displaystyle \tA$, we use interpolation followed by convolution. $\displaystyle \tC$ is first upsampled to a tensor $\displaystyle \tB' \in \R^{h \times w \times c_2}$ if it was downsampled from $\displaystyle \tB$. This transformation is not novel. However, we observe if  $\displaystyle c_1 = c_2$ then we can use the same convolutional layer L (with unit strides) to transform $\displaystyle \tB'$ to $\displaystyle \tA'$. That is, if L does not change the number of feature maps, we can simply use it as a decoding layer instead of adding an additional layer L'.

When L changes the channel dimension, we cannot directly feed $\displaystyle \tB'$ to L and get $\displaystyle \tA'$. However, a layer is just an encapsulation of the underlying operation. To transform $\displaystyle \tA$ to $\displaystyle \tB$, the underlying convolution of L needs a weight tensor $\displaystyle \tW \in \R^{m \times n \times c_1 \times c_2}$. Likewise, to transform $\displaystyle \tB'$ to $\displaystyle \tA'$, the underlying convolution of a layer L' requires a weight tensor $\displaystyle \tW' \in \R^{m \times n \times c_2 \times c_1}$. We find the encoding weight $\displaystyle \tW$ and the decoding weight $\displaystyle \tW'$ are equal in size. This equivalence conveniently makes it possible for us to use L for both encoding and decoding by deriving $\displaystyle \tW'$ from $\displaystyle \tW$.

There are many possible ways to derive $\displaystyle \tW'$ from $\displaystyle \tW$. We found that a simple and efficient method is to permute the dimensions of $\displaystyle \tW$ so that $\displaystyle \tW$ has the same dimensions $\displaystyle \tW'$ requires. In other words, $\displaystyle \tW' \in \R^{h \times w \times c_2 \times c_1}$ can be derived from $\displaystyle \tW \in \R^{h \times w \times c_1 \times c_2}$ by simply swapping the input channel dimension  $\displaystyle c_1$ and the output channel dimension $\displaystyle c_2$. This optimization allows us to realize that a convolutional layer is inherently capable of projecting features to a different dimension in one pass and reversing its effect in the opposite pass. Moreover, this non-destructive derivation of $\displaystyle \tW'$ preserves the inner structure of the original convolutional filters in $\displaystyle \tW$.

To elaborate, let us represent $\displaystyle \tW \in \R^{m \times n \times c_1 \times c_2}$ as a filter matrix $\displaystyle \mW_F \in \R^{c_1 \times c_2}$ whose entries are convolutional filters $\displaystyle F_{i,j} \in \R^{m \times n}$, where $\displaystyle 0 \leq i < c_1$ and $\displaystyle 0 \leq j < c_2$. In the forward pass, each column of filters in $\displaystyle \mW_F$ works as a group to output a single number at every spatial location.

To derive the filter matrix $\displaystyle \mW'_F \in \R^{c_2 \times c_1}$ for the backward pass, we transpose $\displaystyle \mW_F \in \R^{c_1 \times c_2}$ by swapping the input channel dimension and the output channel dimension of $\displaystyle \tW \in \R^{m \times n \times c_1 \times c_2}$. Because each column in $\displaystyle \mW'_F$ was once a row in $\displaystyle \mW_F$, grouping the convolutional filters of $\displaystyle \mW'_F$ into columns is equivalent to grouping the convolutional filters of $\displaystyle \mW$ into rows. This means a convolutional layer can change the number of feature maps in the opposite direction by regrouping its filters and the filters themselves can be kept intact.

\section{Experiments}\label{sec:experiments}
We evaluate the performance of feedbackward decoding on the Cityscapes \citep{DBLP:journals/corr/CordtsORREBFRS16} semantic segmentation benchmark. We chose VGG-16 \citep{Simonyan14c} as the base network due to its straightforward and simple architecture. It consists of 16 trainable layers. The first 13 layers are convolutional and the last 3 layers are fully connected.

We removed the last 3 fully connected layers because they contain nearly 90\% of the parameters but note they may be reused by casting them to convolutional layers with a kernel size of $\displaystyle 1 \times 1$. In place of the fully connected layers we added one additional convolutional layer since that is the minimum requirement to transform the final feature maps to class score maps. For each convolutional layer, we discard the bias and add two batch normalization \citep{DBLP:journals/corr/IoffeS15} operations to handle the shifting - one for the forward pass and one for the backward pass. The activation functions are set to be ReLU \citep{Nair:2010:RLU:3104322.3104425} throughout the network except for the last layer, which is set to be softmax (\citealp[P.~180]{Goodfellow-et-al-2016}). No other layers are added to allow comparison to the original VGG-16 network.

\begin{figure}[p]
\includegraphics{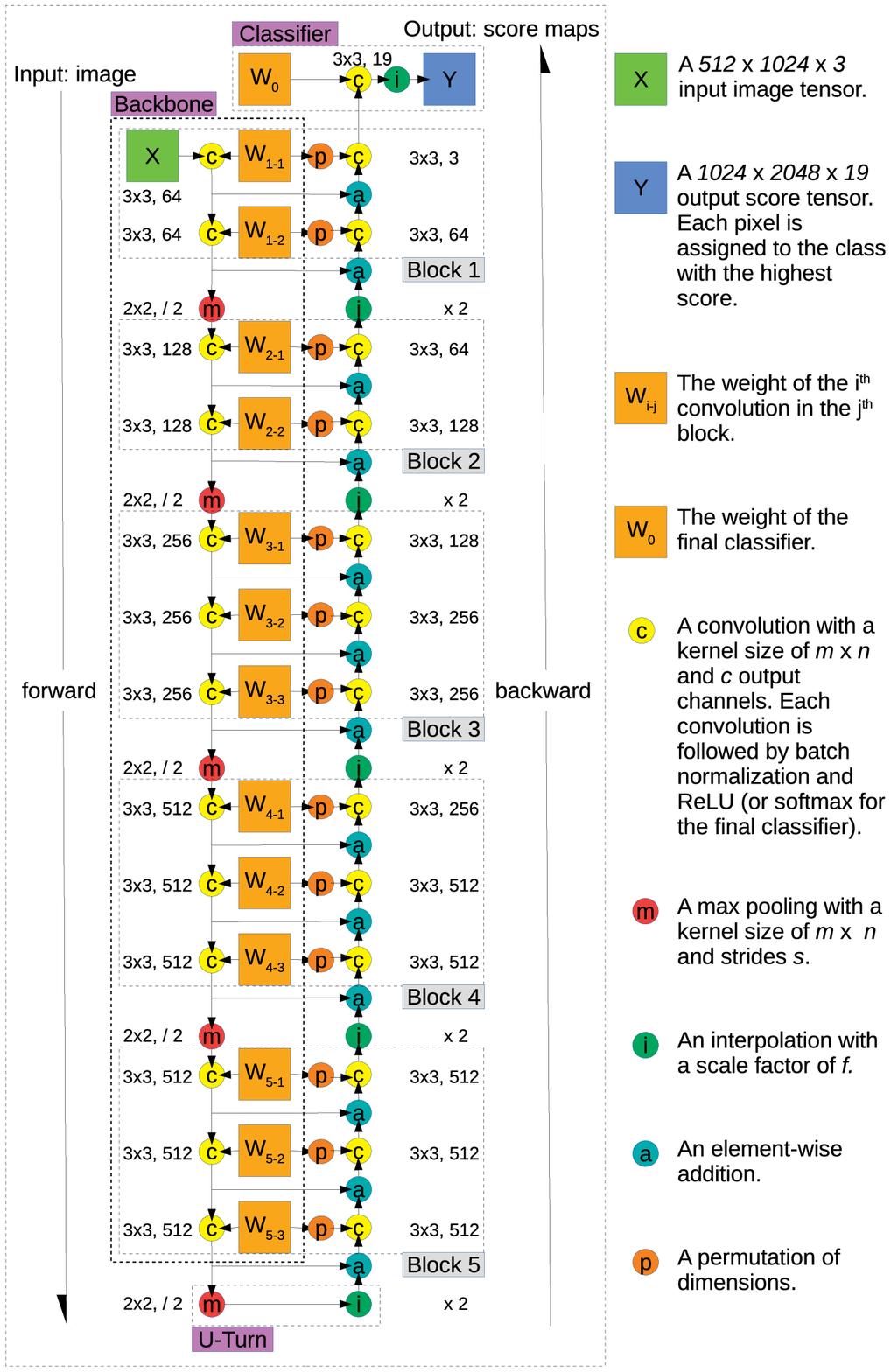}
\caption{VGG-PWP}
\label{fig:vgg_pwp}
\end{figure}

\begin{figure}[p]
\includegraphics{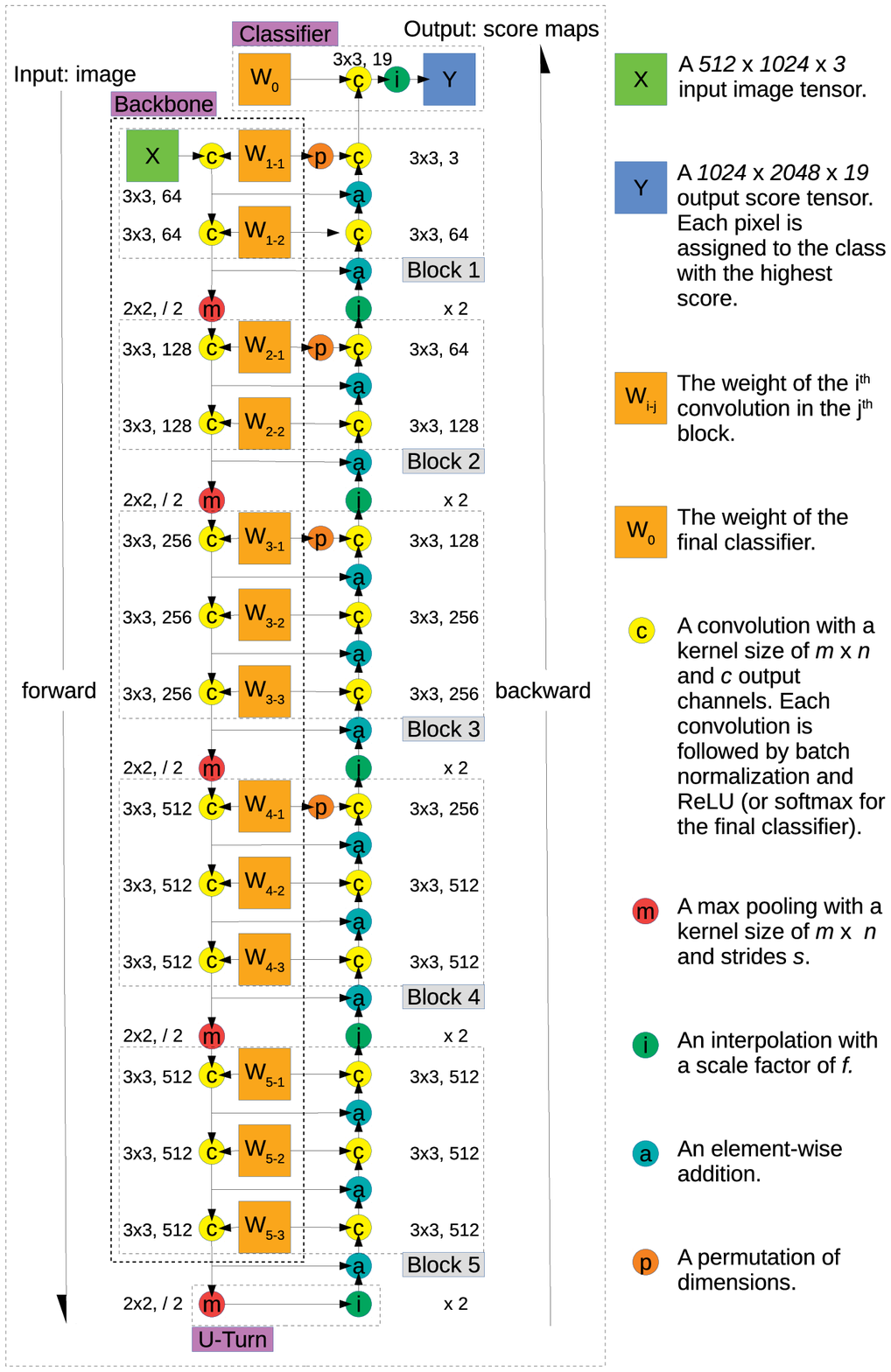}
\caption{VGG-PWN}
\label{fig:vgg_pwn}
\end{figure}

We experimented with two variants of feedbackward decoding VGG, which we call VGG-PWP and VGG-PWN. PWP means the dimensions of pretrained convolutional weights are \textbf{p}ermuted \textbf{w}henever \textbf{p}ossible in the backward pass, as shown in Figure \ref{fig:vgg_pwp}; PWN means they are \textbf{p}ermuted \textbf{w}hen \textbf{n}ecessary, as shown in Figure \ref{fig:vgg_pwn}. The two variants have the exact same layers. The only difference is when the dimensions of weights are permuted in the backward pass.

The adapted VGG networks consist of three parts, fully convolutional layers, a transitional U-turn stage, and a final classifier. The fully convolutional layers are divided into 5 blocks based on the number of output channels. Adjacent blocks are separated by max pooling in the forward pass and interpolation in the backward pass. There are several modes of interpolation. In this work we used the simplest nearest neighbor interpolation.

In the forward pass, information flows from the first layer of the network to the last one. This is the same as all encoder implementations. Once the last layer is reached, a downsampling followed by an upsampling reverses the direction of information flow. Feature maps at depth $\displaystyle d$ in the backward pass are added with the ones at depth $\displaystyle d - 1$ from the forward pass. The only exception is the feature maps at depth $\displaystyle 0$, which are directly fed to the final classifier. The fused feature maps at depth $\displaystyle d$ are then fed to a convolutional layer at depth $\displaystyle d - 1$ in the reverse direction to generate the feedbackward features at depth $\displaystyle d - 1$. If the convolutional layer at depth $\displaystyle d - 1$ changes the channel dimension then the dimensions of its weight need to be permuted as described in Section \ref{sec:methods}. 

In the backward pass, information flows from the last layer to the first layer. It is then sent to the final classification layer. Wherever max pooling was performed in the forward direction, interpolation is performed when processing the backward direction. Additionally, the final score maps are bilinearly interpolated from $\displaystyle 512 \times 1024$ to $\displaystyle 1024 \times 2048$ because the input image is decimated from $\displaystyle 1024 \times 2048$ to $\displaystyle 512 \times 1024$.

There are four evaluation metrics, "IoU class", "iIoU class", "IoU category" and "iIoU category" \citep{DBLP:journals/corr/CordtsORREBFRS16} for the Cityscapes semantic segmentation benchmark. Our training process primarily maximizes the "IoU class" score, which is the intersection over union averaged over the 19 classes. We choose to minimize the loss function proposed in \cite{10.1007/978-3-319-50835-1_22} as it directly approximates the evaluation metric and is more robust to class imbalance. We chose Adam \citep{Kingma2014AdamAM} with the default parameters to be our optimizer and we set the batch size to be 2. To reduce training time, we did not use the additional weakly labeled samples. Additionally random horizontal flipping is the only transformation applied. We train the network on the training set with a constant learning rate of 0.001 and terminate the training process when the performance on the validation set starts to degrade.

\section{Results}\label{sec:results}
\begin{table}[t]
\caption{Cityscapes semantic segmentation results}
\label{cityscapes_results}
\begin{center}
\begin{tabular}{llllll}
\multicolumn{1}{c}{\bf model}
&\multicolumn{1}{c}{\bf IoU class}
&\multicolumn{1}{c}{\bf iIoU class}
&\multicolumn{1}{c}{\bf IoU category}
&\multicolumn{1}{c}{\bf iIoU category}
&\multicolumn{1}{c}{\bf params}
\\ \hline \\
SegNet &57.0 &32.0 &79.1 &61.9 &29.4M \\
DeepLab LargeFOV &64.8 &34.9 &81.3 &58.7 &20.5M \\
VGG-PWP &64.8 &41.9 &86.3 &71.8 &\textbf{14.7M} \\
FCN-8s &65.3 &41.7 &85.7 &70.1 &134.5M \\
\textbf{VGG-PWN} &\textbf{67.3} &\textbf{45.3} &\textbf{88.1} &\textbf{73.5} &\textbf{14.7M} \\
\end{tabular}
\end{center}
\end{table}

In Table \ref{cityscapes_results}, we list the results on the Cityscapes semantic segmentation benchmark from our model and other models from frequently cited papers. To be fair, we only compare our model with models that also use VGG-16 as their base network. This way we can ensure the differences in performance do not come from different choices in base models. The table shows our model significantly outperforms the other VGG-16-based models for every metric. Significantly, our model also has the smallest number of parameters.

Both variants dramatically outperform SegNet while using 50.0\% less parameters. The relative increases in the four scores achieved by VGG-PWP are 13.7\%, 31.0\%, 9.1\% and 16.0\% respectively; The relative increases achieved by VGG-PWN are 18.1\%, 41.6\%, 11.4\% and 18.7\% respectively.

Compared to DeepLab LargeFOV, our networks use 28.3\% less parameters. The PWP variant and DeepLab LargeFOV perform equally well on the "IoU class" metric but the former is significantly better than the latter on the other metrics. The performance of VGG-PWP is 20.1\% better than DeepLab LargeFOV on "iIoU class", 6.2\% better on "IoU category" and 19.4\% better on "iIoU category". The PWN variant also beats DeepLab LargeFOV on these three metrics. The relative improvements are 30.0\%, 8.4\% and 25.2\% respectively. In addition, the "IoU class" score of VGG-PWN is 3.9\% higher than the one of DeepLab LargeFOV.

Our networks are 89.1\% smaller than FCN-8s in model size. Nonetheless, the four scores of VGG-PWN are 3.1\%, 8.6\%, 2.8\% and 4.9\% higher than the respective ones of FCN-8s. Although the "IoU class" score of VGG-PWP is 0.8\% lower than the one of FCN-8s, the other three scores are 0.5\%, 0.7\% and 2.4\% higher.

\section{Discussion}\label{sec:discussion}
 Structurally, SegNet is the closest VGG-16-based model to our adapted VGG-16. It uses only the 13 convolutional layers from VGG-16 for encoding. It also uses a decoder topologically similar to the encoder. The main difference is that the decoder of SegNet is comprised of physical layers while the decoding layers of our model share the encoder layers. In the conventional sense of layers, there are 26 convolutional layers in SegNet and 14 convolutional layers in our model. In \citealp[P.~341]{Goodfellow-et-al-2016}, one way to define the depth of a convolutional neural network is by counting the number of layers that have kernel tensors, or equivalently convolutional layers in this case. By that definition, SegNet is almost twice as deep as our model. However, if we measure the depth of our model not by its number of convolutional layers but by the total number of times information passes through a convolutional layer, then the depth of out model is effectively 27. Therefore, despite having less independent layers, our model is as deep as SegNet. This means our model and SegNet are comparable in terms of model capacity. However, since feedbackward decoding only requires half the number of parameters, our model is much less prone to overfitting (\citealp[P.~112]{Goodfellow-et-al-2016}). It is possible the better generalization capability of feedbackward decoding contributes to our model significantly outperforming SegNet.

DeepLab LargeFOV is one of the first semantic segmentation networks to use atrous convolution. Atrous convolution is now used in almost all state-of-the-art semantic segmentation models. It also uses conditional random fields as a post-processing technique. In addition, it is trained with 19,998 more samples with coarse annotations, whereas our model is only trained with the 2,975 fully labeled samples. Far fewer samples are needed in our training process because the backward pass of our model uses the same weights as the forward pass.

Both FCN-8s and our method consider how convolution can be manipulated to work in both directions. However, the authors of FCN-8s did not use the existing convolutional weights and instead trained new weights.

When a convolutional layer does not change the channel dimension of feature maps, the input and output dimensions of its weight can be either permuted or left as they are in the backward pass. Our experiments showed that better results were achieved by permuting the dimensions only when necessary instead of permuting them whenever possible. As mentioned in section \ref{sec:methods}, convolutional filters are trained to work effectively in teams. A permutation breaks up the grouping and the regrouped filters may not work as effectively as when paired with their original groupings. By permuting the dimensions only when necessary, more beneficial filter groups can be kept together and the overall performance is improved.

\section{Conclusions}\label{sec:conclusions}
Current semantic segmentation models are typically feedforward networks. A feedforward convolutional neural network is a path that has an origin and a destination. A high resolution image leaves the origin and becomes a set of low resolution feature maps at the destination. Since a single convolutional layer can process inputs from both sides, a convolutional neural network can reverse its entrance and exit. When information can flow backward, semantic segmentation no longer requires additional decoding layers. In this paper we show that the best variant of our VGG-16 based model uses 50.0\% less parameters than SegNet, 28.3\% less parameters than DeepLab LargeFOV and 89.1\% less parameters than FCN-8s; and the relative improvements on the "IoU class" metric are 18.1\%, 3.9\% and 3.1\% respectively.
This provides support that semantic segmentation may be better improved by efficiently utilizing existing parameters rather than stacking more layers. 
\bibliography{references}
\bibliographystyle{iclr2019_conference}

\end{document}